%% file: neurips_2026.tex
\documentclass{article}

\usepackage[preprint]{neurips_2026}

\usepackage[utf8]{inputenc}
\usepackage[T1]{fontenc}
\usepackage{microtype}
\usepackage{hyperref}
\usepackage{url}
\usepackage{booktabs}
\usepackage{colortbl}
\usepackage{amsmath}
\usepackage{amssymb}
\usepackage{amsfonts}
\usepackage{nicefrac}
\usepackage{graphicx}
\usepackage{xcolor}
\usepackage{multirow}
\usepackage{subcaption}
\usepackage{enumitem}
\usepackage{algorithm}
\usepackage{algorithmic}
\usepackage{wrapfig}
\usepackage{comment}

\definecolor{darkblue}{rgb}{0, 0, 0.5}
\hypersetup{colorlinks=true, citecolor=darkblue, linkcolor=darkblue, urlcolor=darkblue}

\newcommand{\ours}{\textsc{RELEX}}
\newcommand{\vr}{V_r}
\newcommand{\Ct}{\mathbf{C}}
\newcommand{\dt}{\Delta\theta}
\newcommand{\R}{\mathbb{R}}

\newcommand{\Tobs}{T_{\text{cut}}}

\usepackage{xspace} 

\newcommand{\ie}{{\sl i.e.}}
\newcommand{\eg}{{\sl e.g.}}

\title{You Only Need Minimal RLVR Training: Extrapolating LLMs via Rank-1 Trajectories}

\author{Zhepei Wei$^\dagger$ \quad Xinyu Zhu$^\dagger$ \quad Wei-Lin Chen$^\dagger$ \quad Chengsong Huang$^\ddagger$ \\\quad {\bf Jiaxin Huang}$^\ddagger$ \quad {\bf Yu Meng}$^\dagger$\\
$^\dagger$University of Virginia \quad $^\ddagger$Washington University in St. Louis\\
\texttt{\{zhepei.wei,xinyuzhu,wlchen,yumeng5\}@virginia.edu}\\\texttt{\{chengsong,jiaxinh\}@wustl.edu}
}

\begin{document}

\maketitle

\input{tex/0-abs}
\input{tex/1-intro}
\input{tex/2-background}
\input{tex/3-method}
\input{tex/4-exp}

\input{tex/5-related}
\input{tex/6-discussion}
\input{tex/7-conclusion}

\bibliography{references}
\bibliographystyle{plainnat}

\newpage
\appendix
\input{tex/appendix}

\end{document}

%% file: tex/0-abs.tex
\begin{abstract}
Reinforcement learning with verifiable rewards (RLVR) has become a dominant paradigm for improving reasoning in large language models (LLMs), yet the underlying geometry of the resulting parameter trajectories remains underexplored.
In this work, we demonstrate that RLVR weight trajectories are \emph{extremely low-rank} and \emph{highly predictable}. Specifically, we find that the majority of downstream performance gains are captured by a rank-1 approximation of the parameter deltas, where the magnitude of this projection evolves near-linearly with training steps.
Motivated by this, we propose a simple and compute-efficient method \textsc{RELEX} ({\bf RE}inforcement {\bf L}earning {\bf EX}trapolation), which estimates the rank-1 subspace from a short observation window and extrapolates future checkpoints via linear regression, with no learned model required.
Across three models (\ie, Qwen2.5-Math-1.5B, Qwen3-4B-Base, and Qwen3-8B-Base), \textsc{RELEX} produces checkpoints that match or exceed RLVR performance on both in-domain and out-of-domain benchmarks, requiring as few as 15\% steps of full RLVR training.
Remarkably, \textsc{RELEX} is able to extrapolate far beyond the observation window at no training cost, predicting checkpoints up to $10\sim20\times$ beyond the observed prefix with continued improvement (\eg, observe only the first 50 steps and extrapolate to 1000 steps).
Our ablation analysis confirms the minimalist sufficiency of \textsc{RELEX}: neither increasing the subspace rank nor employing non-linear modeling yields further gains in extrapolation.
Finally, we show that RELEX’s success stems from a ``denoising'' effect: by projecting updates onto the rank-1 subspace, the model discards stochastic optimization noise that would otherwise degrade performance during extrapolation.
Our code is available at \url{https://github.com/weizhepei/RELEX}.
\end{abstract}

%% file: tex/1-intro.tex
\section{Introduction}
\label{sec:intro}

Reinforcement learning with verifiable rewards (RLVR) has become a central technique for unlocking reasoning capabilities in large language models~\citep{lambert2025tulu,guo2025deepseek}. A typical RLVR pipeline trains an LLM over massive optimization steps using algorithms such as Group Relative Policy Optimization~(GRPO;~\citealp{shao2024deepseekmath}), producing a trajectory of checkpoints that progressively improve on target tasks. However, this process is computationally expensive, often requiring days of GPU time even for moderately sized models~\citep{yang2025qwen3,olmo2025olmo}, and the cost scales directly with the number of training steps~\citep{liu2025prorl}.

Prior works~\citep{yue2025does,zhu2025the} show that RLVR appears to operate less by teaching entirely new capabilities from scratch than by eliciting and amplifying behaviors already latent in the pretrained model---it tends to increase the likelihood of successful reasoning traces while suppressing incorrect modes. Recent analyses further reveal that RLVR updates are highly structured~\citep{wang2026linearity,zhu2025rlvr}, suggesting that the update directions can matter more than magnitude~\citep{huang2026beyond} and that RLVR may modify only sparse or low-dimensional subsets of parameters~\citep{mukherjee2025subnetworks,shenfeld2026rls}. 
This raises a natural question: \emph{can we predict where RLVR training is heading from its early dynamics?} We hypothesize that the trajectory of RLVR updates follows a structured pattern, where future checkpoints could be predicted from a short prefix (\eg, the first 15\% of steps), while achieving the same level of performance as the fully trained model.

In this work, we study weight update trajectories during RLVR training and reveal two key structural findings.
First, \textbf{RLVR updates are low-rank}: denote $\theta_0$ as the weight of a base model and $\theta_t$ as the weight of its RLVR-ed counterpart trained for $t$ steps. By computing parameter deltas $\dt_t = \theta_t - \theta_0$ and applying singular value decomposition (SVD), we find that a single dominant direction (rank-1) per weight tensor captures most downstream-relevant parameter change.
Specifically, we find that the rank-1 reconstructed checkpoint closely matches the oracle RLVR checkpoint across training steps and model families.
Second, \textbf{the rank-1 coefficient evolves near-linearly}: projecting each tensor's trajectory onto its dominant singular vector yields a scalar sequence $c_t$ that is well-approximated by a linear function of training step, with $R^2 > 0.98$ ($R^2 =1$ means perfect fit) for most tensors (\S\ref{sec:preliminary}).

Motivated by these findings, we introduce \textsc{RELEX}, a simple training-free method that first estimates the rank-1 subspace from the first $\Tobs$ steps via SVD, then fits a line to the projected coefficients, and finally extrapolates future checkpoints via linear regression (\S\ref{sec:method}). No learned model is required, and once the subspace is estimated, predicting any future checkpoint is training-free.
For instance, with 15--20\% of RLVR's training cost, \textsc{RELEX} matches or exceeds GRPO on Qwen2.5-Math-1.5B (71.6\% vs.\ 71.5\%), Qwen3-4B-Base (85.6\% vs.\ 85.5\%), and Qwen3-8B-Base (87.4\% vs. 88.5\%) on the in-domain MATH benchmark, while also outperforming RLVR across five out-of-domain (OOD) benchmarks on average.
Interestingly, our analysis shows that the dominant rank-1 component explains most update variance, while higher-rank components capture trivial dynamics, suggesting that rank-1 projection acts as a spectral denoiser, preserving the stable task-relevant signal while discarding stochastic optimization noise (\S\ref{sec:ablation}).
We summarize our contributions as follows.

\begin{figure}[t!]
\centering
\includegraphics[width=\textwidth]{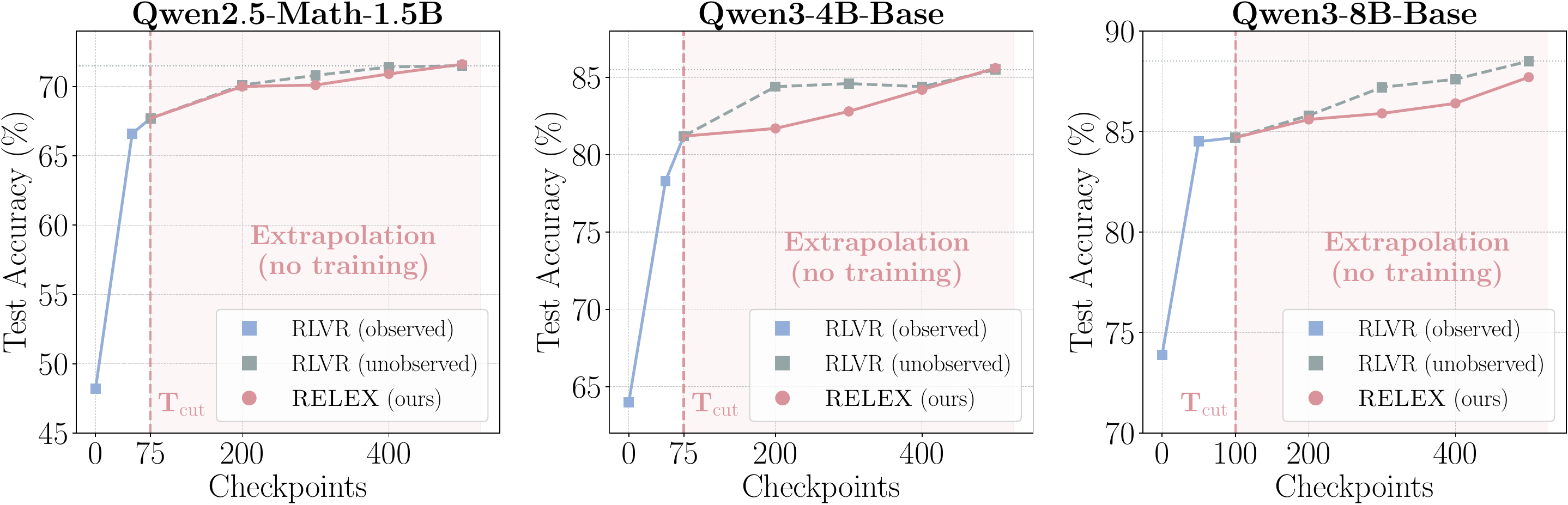}
\vspace{-1.5em}
\caption{\textbf{RELEX extrapolates checkpoints that match full RLVR performance based only on early training dynamics, without further training.} \textsc{RELEX} estimates the rank-1 update subspace from the observed RLVR prefix (up to $\Tobs$) and extrapolates future checkpoints at no training cost, matching or exceeding the RLVR checkpoints on the MATH test set across three models.
}
\label{fig:teaser}
\vspace{-2em}
\end{figure}

\begin{itemize}[leftmargin=1.5em,topsep=0.2em,itemsep=0.1em]
  \item We empirically demonstrate that RLVR weight update trajectories are extremely low-rank and near-linear across training steps: rank-1 SVD captures the dominant update direction, with rank-1 reconstructed checkpoints closely matching RLVR checkpoints across training steps (\S\ref{sec:preliminary}).
  \item We propose \textsc{RELEX}, a simple training-free method that predicts future RLVR checkpoints via rank-1 SVD projection and linear extrapolation, with no learned model required (\S\ref{sec:method}). Empirical results show that \textsc{RELEX} with as few as 15\% of training cost can match and often exceed full RLVR on both in-domain and OOD math benchmarks across three backbone models.
  \item Our analysis shows that neither increasing the subspace rank nor employing non-linear modeling yields further gains in extrapolation, confirming the minimalist sufficiency of \textsc{RELEX} (\S\ref{sec:ablation}).
\end{itemize}

%% file: tex/2-background.tex
\section{Background}
\label{sec:background}

\subsection{Reinforcement Learning with Verifiable Rewards}

RLVR algorithms train an LLM policy $\pi_\theta$ to maximize rewards that can be programmatically verified, such as mathematical solution correctness~\citep{guo2025deepseek}. 
In this work, we adopt Group Relative Policy Optimization (GRPO;~\citealp{shao2024deepseekmath}) as the RL algorithm. For each prompt, it samples multiple responses from a snapshot policy, scores them with the verifier, and updates $\pi_\theta$ via a token-level clipped objective regularized by a KL penalty toward a reference policy. We refer to~\citet{shao2024deepseekmath} for more details.
In practice, RLVR runs for a massive number of optimization steps, producing a trajectory of checkpoints that improve on the target task until plateauing.

\subsection{SVD of Parameter Trajectories}
\label{sec:svd_background}
\begin{wrapfigure}[22]{r}{0.52\textwidth}
\vspace{-2.2em}
\begin{minipage}{0.52\textwidth}
\begin{algorithm}[H]
\caption{Low-rank SVD Reconstruction of RLVR Weight Trajectories}
\label{alg:decompose}
\begin{algorithmic}[1]
\REQUIRE Checkpoints $\{\theta_0, \theta_1, \ldots, \theta_{\Tobs}\}$, rank $r$
\ENSURE Reconstructed checkpoints $\{\hat{\theta}_t\}_{t=1}^{\Tobs}$
\STATE {\color{blue}// Step 1: Construct trajectory matrices}
\STATE \textbf{For} each parameter tensor $W^{(\ell)}$:
\STATE \quad $\dt_t^{(\ell)} \gets W_t^{(\ell)} - W_0^{(\ell)}$, $t = 1, \ldots, \Tobs$
\STATE \quad $\mathbf{m}_t \gets \text{flatten}(\dt_t^{(\ell)}) \in \R^{d_\ell}$
\STATE $\mathbf{M}^{(\ell)} \gets \text{stack}(\mathbf{m}_1, \ldots, \mathbf{m}_{\Tobs})$
\STATE {\color{blue}// Step 2: SVD and truncation}
\STATE $\mathbf{U}, \boldsymbol{\Sigma}, \mathbf{V}^\top \gets \text{SVD}(\mathbf{M}^{(\ell)})$
\STATE $\vr^{(\ell)} \gets \mathbf{V}^\top[:r, :]$ \hfill $\triangleright$ Top-$r$ directions
\STATE {\color{blue}// Step 3: Project to low-rank space}
\STATE $\Ct_r^{(\ell)} \gets \mathbf{U}[:,:r]\, \boldsymbol{\Sigma}[:r,:r]$ \hfill $\triangleright$ Coefficients
\STATE {\color{blue}// Step 4: Reconstruct checkpoints}
\STATE \textbf{For} $t = 1, \ldots, \Tobs$:
\STATE \quad $\hat{W}_t^{(\ell)} \gets W_0^{(\ell)} + \Ct_r^{(\ell)}[t] \cdot \vr^{(\ell)}$
\RETURN $\{\hat{\theta}_t\}_{t=1}^{\Tobs}$ where $\hat{\theta}_t = \{\hat{W}_t^{(\ell)}\}_\ell$
\end{algorithmic}
\end{algorithm}
\end{minipage}
\end{wrapfigure}
Given a sequence of RLVR checkpoints $\{\theta_0, \theta_1, \ldots, \theta_T\}$, we compute parameter deltas $\dt_t = \theta_t - \theta_0$ relative to the base model.
For each parameter tensor (\eg, an attention weight matrix $W \in \R^{m \times n}$), we flatten and stack the deltas into a trajectory matrix $\mathbf{M} \in \R^{T \times mn}$ whose $t$-th row is $\dt_t$.

The compact SVD $\mathbf{M} = \mathbf{U} \boldsymbol{\Sigma} \mathbf{V}^\top$ decomposes this trajectory into a \emph{subspace} $\mathbf{V}$ (directions along which parameters change) and \emph{coefficients} $\Ct = \mathbf{U}\boldsymbol{\Sigma}$ (temporal dynamics within that subspace). A rank-$r$ truncation gives:
\begin{equation*}
    \hat{\theta}_t = \theta_0 + \Ct_r[t] \cdot \vr,
\end{equation*}
where $\Ct_r[t] \in \R^r$ is the $t$-th row of the truncated coefficient matrix and $\vr \in \R^{r \times mn}$ contains the top-$r$ right singular vectors. This factorization cleanly separates \emph{where} parameters move (subspace~$\vr$) from \emph{when and how much} they move (coefficients~$\Ct_r$), enabling independent analysis and prediction of each component.

%% file: tex/3-method.tex
\section{Method}

\begin{figure}[t!]
\centering
\includegraphics[width=0.95\textwidth]{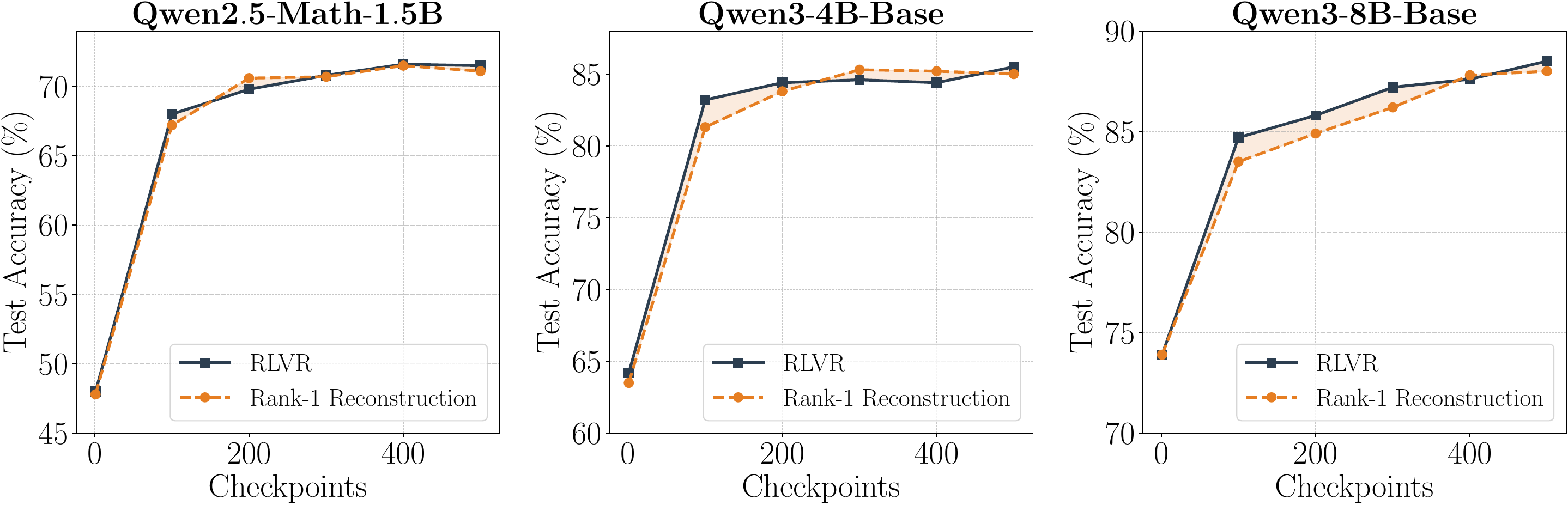}
\vspace{-0.5em}
\caption{\textbf{Rank-1 SVD reconstruction recovers RLVR checkpoints across models.} The rank-1 reconstructed checkpoints preserve most downstream performance on MATH, suggesting that a single dominant direction captures the task-relevant component of RLVR updates.
}
\label{fig:reconstruction}
\vspace{-1em}
\end{figure}

\subsection{RLVR Weight Trajectories Are Extremely Low-Rank and Predictable}
\label{sec:preliminary}

Before developing our extrapolation method, we examine whether RLVR weight updates exhibit structured patterns that make prediction tractable. We compute parameter deltas $\dt_t = \theta_t - \theta_0$ for each of the 500 RLVR training steps on Qwen2.5-Math-1.5B, perform per-tensor SVD on the resulting trajectory matrices (Algorithm~\ref{alg:decompose}), and observe two insightful empirical findings.

\begin{figure}[t!]
\centering
\includegraphics[width=0.95\textwidth]{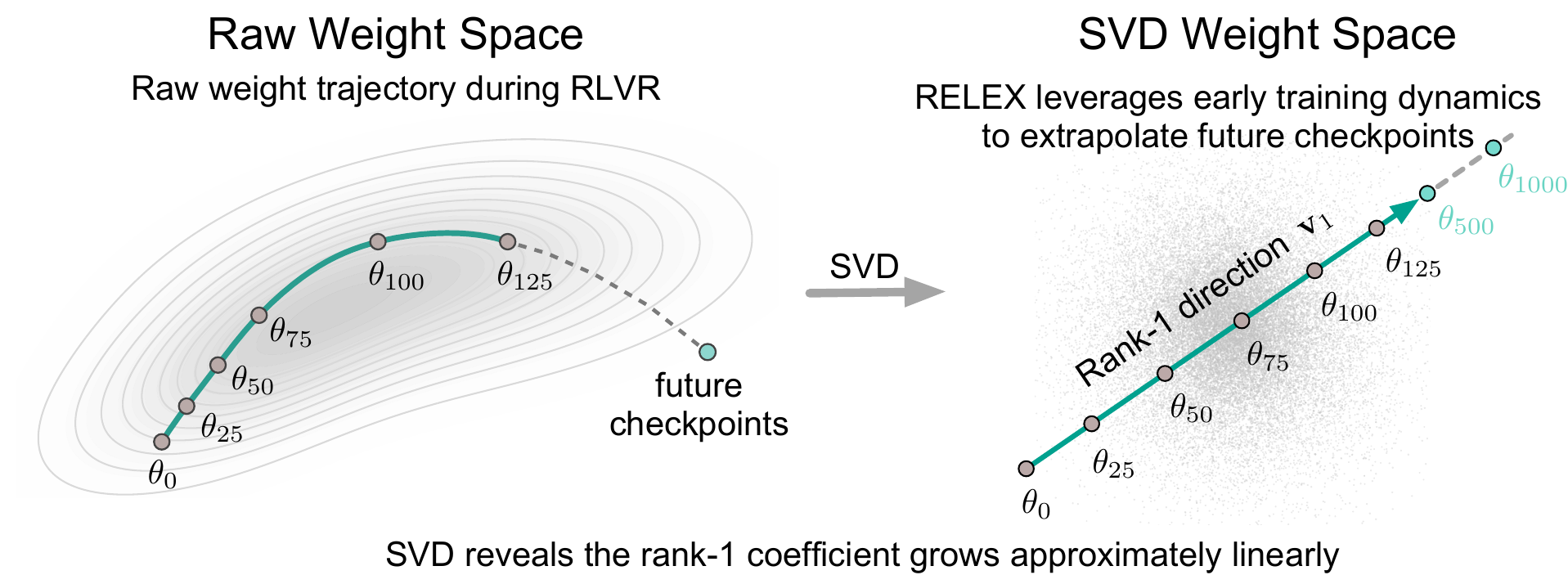}
\caption{\textbf{From raw RLVR trajectories to rank-1 extrapolation.} \emph{Left:} RLVR checkpoints form a curved path in raw weight space, making future checkpoints hard to predict directly. \emph{Right:} after SVD, the dominant direction $\mathbf{v}_1$ captures the main update, and the corresponding scalar coefficient grows approximately linearly with training step. \textsc{RELEX} uses the observed prefix $\theta_{\le \Tobs=125}$ to estimate $\mathbf{v}_1$, fits this rank-1 coefficient trajectory, and extrapolates along the same direction to predict future checkpoints such as $\theta_{500}$ or $\theta_{1000}$ without additional RLVR training.}
\label{fig:method}
\vspace{-1em}
\end{figure}

\begin{figure}[t!]
\centering
\includegraphics[width=0.98\textwidth]{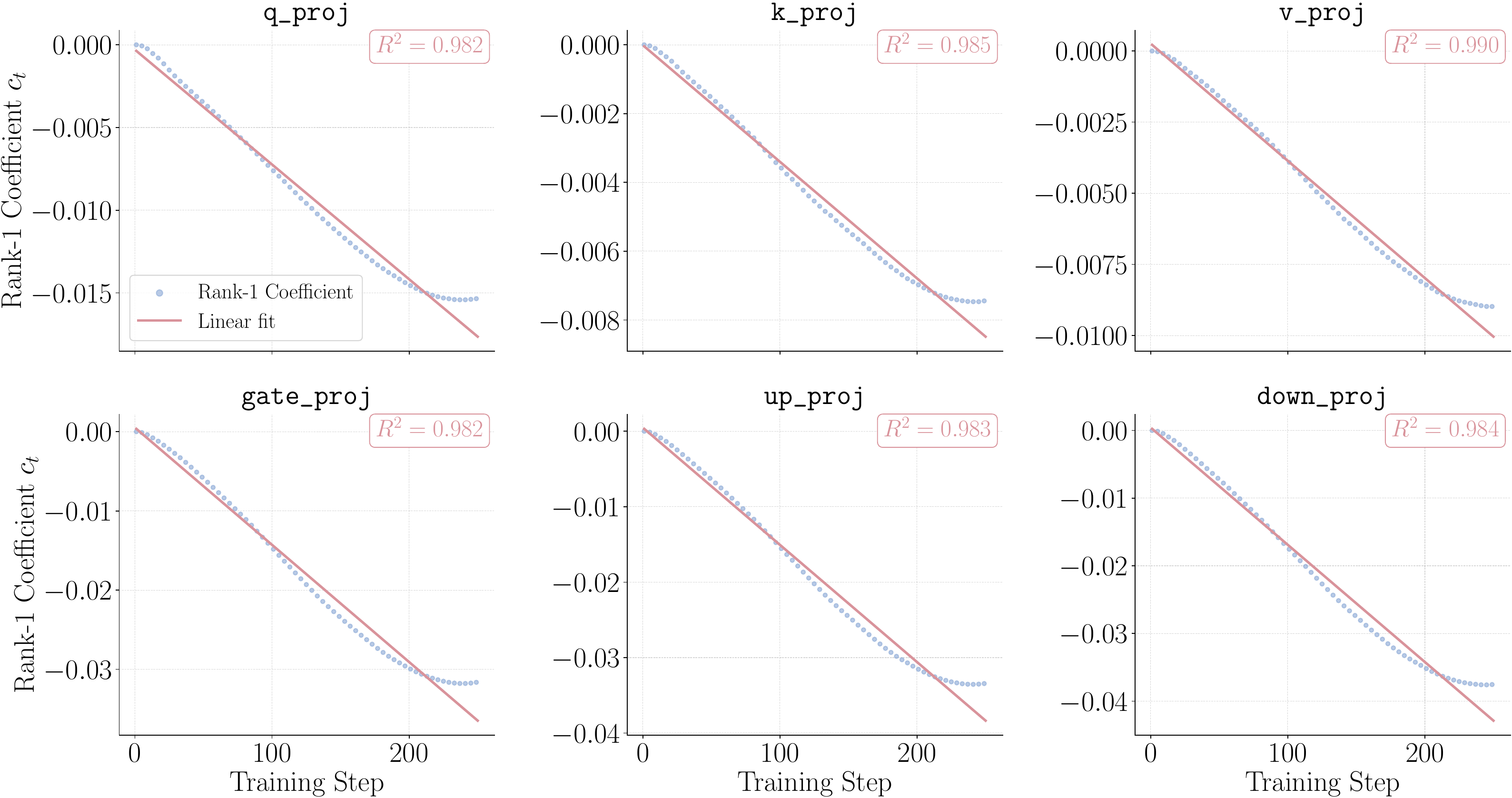}
\caption{\textbf{Rank-1 SVD coefficients evolve nearly linearly.} Rank-1 coefficients $c_t$ (blue dots) and linear fits (pink) for representative modules of Qwen2.5-Math-1.5B.
}
\vspace{-1em}
\label{fig:coeff_linearity}
\end{figure}

{\bf Finding 1: RLVR updates are low-rank.}
Figure~\ref{fig:reconstruction} shows that across all three models, rank-1 SVD reconstruction closely tracks the RLVR trajectory: replacing each trained tensor with its rank-1 approximation preserves nearly all of the downstream MATH accuracy gain over the base model. Although weight tensors live in a high-dimensional space and could in principle move along many independent components, a single component per tensor accounts for nearly all task-relevant change.

{\bf Finding 2: The rank-1 coefficient evolves linearly in training step.}
The temporal dynamics within the rank-1 subspace are surprisingly simple. We project each observed delta onto $\mathbf{v}_1$ to obtain a trajectory of scalar coefficient  $c_t$, then fit $c(t) = at + b$ via least squares.
Figure~\ref{fig:coeff_linearity} plots this fit on representative modules. Across the RLVR training trajectory, the coefficient closely tracks a single straight line, with $R^2 > 0.98$ across most tensors, indicating the linearity of rank-1 coefficients.

{\bf From structure to prediction.}
Together, these two findings reduce the prediction of RLVR checkpoints into a straightforward two-step process: (1) estimating the rank-1 direction $\mathbf{v}_1$ from the observed prefix via SVD and (2)~extrapolating the scalar coefficient $c_{T}$ of the target step $T$ via a linear fit.
Figure~\ref{fig:method} illustrates the core intuition, and \textsc{RELEX} (\S\ref{sec:method}) is the direct realization of it.

\subsection{\textsc{RELEX}: Predicting RLVR Checkpoints via Low-Rank Extrapolation}
\label{sec:method}
As shown in Algorithm~\ref{alg:rlex}, given the first $\Tobs$ RLVR checkpoints, \textsc{RELEX} predicts future checkpoints via three steps: (1) rank-1 subspace estimation, (2) linear coefficient extrapolation, and (3) predicting future weights.

\begin{wrapfigure}[21]{r}{0.54\textwidth}
\vspace{-2em}
\begin{minipage}{0.54\textwidth}
\begin{algorithm}[H]
\caption{\textsc{RELEX}: RLVR Extrapolation}
\label{alg:rlex}
\begin{algorithmic}[1]
\REQUIRE Base checkpoint $\theta_0$, RLVR checkpoints $\theta_1, \ldots, \theta_{\Tobs}$, target step $T$
\ENSURE Predicted checkpoint $\hat{\theta}_T$
\STATE \textbf{For} each weight tensor $W^{(\ell)}$:
\STATE {\color{blue}// Step 1: Rank-1 subspace estimation}
\STATE \quad $\dt_t^{(\ell)} \gets W_t^{(\ell)} - W_0^{(\ell)}$, \quad $t = 1, \ldots, \Tobs$
\STATE \quad $\mathbf{m}_t \gets \text{flatten}(\dt_t^{(\ell)})$
\STATE \quad $\mathbf{M}^{(\ell)} \gets \text{stack}(\mathbf{m}_1, \ldots, \mathbf{m}_{\Tobs})$
\STATE \quad $\mathbf{U}, \boldsymbol{\Sigma}, \mathbf{V}^\top \gets \text{SVD}(\mathbf{M}^{(\ell)})$
\STATE \quad $\mathbf{v}_1^{(\ell)} \gets \mathbf{V}^\top[0, :]$ \hfill $\triangleright$ Top-1 singular vector
\STATE {\color{blue}// Step 2: Linear coefficient extrapolation}
\STATE \quad $\Ct_1^{(\ell)} \gets \mathbf{U}[:,0]\, \boldsymbol{\Sigma}[0,0]$ \hfill $\triangleright$ Rank-1 coefficients
\STATE \quad $a^{(\ell)}, b^{(\ell)} \gets \text{LinearFit}(\Ct_1^{(\ell)})$
\STATE \quad $\hat{c}_T^{(\ell)} \gets a^{(\ell)} \cdot T + b^{(\ell)}$
\STATE {\color{blue}// Step 3: Predict future weights}
\STATE \quad $\hat{W}_T^{(\ell)} \gets W_0^{(\ell)} + \hat{c}_T^{(\ell)} \cdot \mathbf{v}_1^{(\ell)}$
\RETURN $\hat{\theta}_T = \{\hat{W}_T^{(\ell)}\}_\ell$
\end{algorithmic}
\end{algorithm}
\end{minipage}
\end{wrapfigure}
\paragraph{Step 1: Rank-1 subspace estimation.}
For each weight tensor $W^{(\ell)}$, we compute parameter deltas $\dt_t^{(\ell)} = W_t^{(\ell)} - W_0^{(\ell)}$ for $t = 1, \ldots, \Tobs$ and stack their vectorized forms into a trajectory matrix $\mathbf{M}^{(\ell)} \in \R^{\Tobs \times d}$. We extract the top right singular vector $\mathbf{v}_1^{(\ell)}$ via truncated SVD. This vector defines the dominant direction of parameter change across the observed training window.

\paragraph{Step 2: Linear coefficient extrapolation.} We project each observed delta onto the rank-1 direction to obtain a scalar coefficient trajectory $\Ct_1^{(\ell)} = [c_1^{(\ell)}, \ldots, c_{\Tobs}^{(\ell)}]$ where $c_t^{(\ell)} = \langle \text{flatten}(\dt_t^{(\ell)}),\, \mathbf{v}_1^{(\ell)} \rangle$. We then fit a linear function $c(t) = at + b$ with slope $a^{(\ell)} = \mathrm{Cov}(t, c_t^{(\ell)}) / \mathrm{Var}(t)$ and intercept $b^{(\ell)} = \bar{c}^{(\ell)} - a^{(\ell)}\bar{t}$, and extrapolate to the target step $T$ as $\hat{c}_T^{(\ell)} = a^{(\ell)} T + b^{(\ell)}$. The justification for this step is our empirical finding (\S\ref{sec:preliminary}) that $c_t^{(\ell)}$ is well-approximated by a linear function with $R^2 > 0.98$ for the vast majority of tensors.
\paragraph{Step 3: Predicting future weights.} We reconstruct the predicted weight tensor as $\hat{W}_T^{(\ell)} = W_0^{(\ell)} + \hat{c}_T^{(\ell)} \cdot \mathbf{v}_1^{(\ell)}$, adding the predicted delta back to the base weights. Assembling predictions across all tensors yields the full predicted checkpoint $\hat{\theta}_T$.
\paragraph{Zero training cost.} Notably, \textsc{RELEX} only requires one truncated SVD per tensor (retaining only the top singular vector) plus a two-parameter least-squares fit---both closed-form and negligible in cost relative to RLVR training itself. The method has no learnable parameters and requires no additional RLVR training beyond the $\Tobs$ observation training window.

%% file: tex/4-exp.tex
\section{Experiments}
\label{sec:experiments}

\subsection{Experimental Setup}

\paragraph{RLVR training and evaluation.} We study RLVR weight trajectories on three models, including Qwen2.5-Math-1.5B~\citep{yang2024qwen2}, Qwen3-4B-Base~\citep{yang2025qwen3}, and Qwen3-8B-Base. All models are trained with GRPO~\citep{shao2024deepseekmath} on MATH~\citep{hendrycks2021math} until they plateau with a total of 500 training steps, with checkpoints saved at every step.
We evaluate on both the in-domain MATH benchmark and five out-of-distribution (OOD) benchmarks: AIME 2025~\citep{dekoninck2026matharena}, AIME 2026, HMMT 2025~\citep{dekoninck2026matharena}, OlympiadBench~\citep{he2024olympiadbench}, and AMC 2023.

\paragraph{Baselines.} We compare \textsc{RELEX} against the following baselines. \textbf{Base} is the pretrained model before any RLVR fine-tuning, serving as a lower bound. \textbf{RLVR} denotes the actual RLVR training checkpoints, which are the target to approximate.
\textbf{ExPO}~\citep{zheng2025expo} amplifies the weight delta from an initial checkpoint to a partially trained checkpoint, using a fixed scalar. \textbf{AlphaRL}~\citep{cai2026on} computes a rank-1 SVD independently at each early checkpoint and uses a PLS regression over these per-checkpoint decompositions to predict a single dominant update vector.
\textbf{Weight Extrap.}~\citep{wang2026linearity} linearly interpolates between two arbitrary checkpoints in raw weight space, without any SVD decomposition. \textbf{Logits Extrap.}~\citep{wang2026linearity} applies the same two-endpoint linear extrapolation in output-logit space at inference time, leaving model weights unchanged. 
Additional implementation details and discussion on baselines are provided in Appendix~\ref{app:implementation}.

\subsection{Results}
\label{sec:results}

\begin{table}[t]
\centering
\small
\caption{Main results across in-domain and out-of-domain benchmarks. Extrapolation rows are shaded; \textbf{Bold} marks the best value per column among extrapolation methods. Base and RLVR rows are shown for reference and are not directly comparable to the extrapolation methods.}\label{tab:main}
\resizebox{\textwidth}{!}{%
\begin{tabular}{lcccccccc}
\toprule
\textbf{Method} & \textbf{MATH} & \textbf{AIME25} & \textbf{AIME26} & \textbf{HMMT25} & \textbf{OlympBench} & \textbf{AMC23} & \textbf{Avg.} & \textbf{Cost} \\
\midrule
\textit{Qwen2.5-Math-1.5B} \\
\quad Base              & 48.2 &  4.2 &  5.0 &  0.8 & 23.2 & 33.1 & 19.1 &   0\% \\
\quad RLVR              & 71.5 &  4.6 &  7.9 &  0.4 & 31.5 & 54.4 & 28.4 & 100\% \\
\rowcolor{gray!15} \quad ExPO           & 67.7 &  6.7 &  8.8 &  0.4 & 29.5 & 50.3 & 27.2 &  15\% \\
\rowcolor{gray!15} \quad AlphaRL       & 67.3 &  4.2 &  5.8 &  1.3 & 28.4 & 50.6 & 26.3 &  15\% \\
\rowcolor{gray!15} \quad Logits Extrap. & 64.9 &  3.8 &  7.9 &  0.4 & 28.2 & 44.8 & 25.0 &  15\% \\
\rowcolor{gray!15} \quad Weight Extrap. & 70.4 & \textbf{9.2} &  7.5 &  0.0 & 30.6 & 52.2 & 28.3 &  15\% \\
\rowcolor{gray!15} \quad {\bf \ours{}}  & \textbf{71.6} &  8.8 & \textbf{10.0} & \textbf{2.1} & \textbf{31.4} & \textbf{56.2} & \textbf{30.0} &  15\% \\
\midrule
\textit{Qwen3-4B-Base} \\
\quad Base              & 64.0 &  7.9 &  8.8 &  0.8 & 31.6 & 43.8 & 26.2 &   0\% \\
\quad RLVR              & 85.5 & 23.8 & 23.8 & 10.0 & 46.6 & 64.1 & 42.3 & 100\% \\
\rowcolor{gray!15} \quad ExPO           & 82.7 & 21.7 & 20.0 &  9.6 & 44.1 & 60.9 & 39.8 &  15\% \\
\rowcolor{gray!15} \quad AlphaRL       & 80.7 & 17.9 & 16.3 &  7.5 & 42.0 & 55.3 & 36.6 &  15\% \\
\rowcolor{gray!15} \quad Logits Extrap. & 79.2 & 14.2 & \textbf{20.8} &  6.7 & 40.1 & 59.1 & 36.7 &  15\% \\
\rowcolor{gray!15} \quad Weight Extrap. & 82.8 & \textbf{25.8} & \textbf{20.8} & 10.4 & 42.9 & 58.8 & 40.3 &  15\% \\
\rowcolor{gray!15} \quad {\bf \ours{}}  & \textbf{85.6} & 23.8 & 19.2 & \textbf{14.6} & \textbf{47.4} & \textbf{67.2} & \textbf{43.0} &  15\% \\
\midrule
\textit{Qwen3-8B-Base} \\
\quad Base              & 73.9 & 10.0 &  7.1 &  2.9 & 36.9 & 53.8 & 30.8 &   0\% \\
\quad RLVR              & 88.5 & 29.2 & 25.4 & 16.3 & 49.5 & 73.8 & 47.1 & 100\% \\
\rowcolor{gray!15} \quad ExPO           & 85.5 & 18.3 & 17.9 & 13.3 & 47.1 & 67.8 & 41.7 &  20\% \\
\rowcolor{gray!15} \quad AlphaRL       & 85.0 & 21.3 & 17.9 & 11.3 & 46.9 & 66.9 & 41.5 &  20\% \\
\rowcolor{gray!15} \quad Logits Extrap. & 81.8 & 19.6 & 20.0 &  8.3 & 46.0 & 68.4 & 40.7 &  20\% \\
\rowcolor{gray!15} \quad Weight Extrap. & 87.2 & 18.8 & 23.3 & \textbf{16.3} & 48.5 & 68.1 & 43.7 &  20\% \\
\rowcolor{gray!15} \quad {\bf \ours{}}  & \textbf{87.4} & \textbf{27.5} & \textbf{24.6} & 15.4 & \textbf{49.6} & \textbf{72.8} & \textbf{46.2} &  20\% \\
\bottomrule
\end{tabular}%
}
\end{table}

\paragraph{\textsc{RELEX} matches full RLVR with 80\% less training cost and generalizes well.} Table~\ref{tab:main} reports the main comparison under matched training costs for extrapolation methods, along with the comparison with base and full RLVR. On in-domain MATH, \textsc{RELEX} matches or slightly exceeds RLVR on the two smaller models---\textbf{71.6\% vs.\ 71.5\%} on Qwen2.5-Math-1.5B and \textbf{85.6\% vs.\ 85.5\%} on Qwen3-4B-Base, and stays within $1.1$\% on Qwen3-8B-Base ($87.4$\% vs.\ $88.5$\%).
On out-of-distribution (OOD) competitions, \textsc{RELEX} outperforms RLVR on 4 of 5 benchmarks for Qwen2.5 (AIME25, AIME26, HMMT25, AMC23) and on 3 of 5 for Qwen3-4B (HMMT25, OlympBench, AMC23). On Qwen3-8B-Base, the overall averages are within $0.9$\% ($46.2$\% vs.\ $47.1$\%), with \textsc{RELEX} still winning OlympBench. Across all three models, \textsc{RELEX} closely recovers full RLVR-level accuracy on the in-domain MATH benchmark and matches or even improves OOD generalization, while paying only $15$--$20\%$ of the RLVR training cost. Particularly, the OOD trend suggests that {\bf \textsc{RELEX}-extrapolated checkpoints capture transferable reasoning gains rather than merely memorizing the MATH training distribution.}

\paragraph{\textsc{RELEX} dominates the other extrapolation baselines at the same compute budget.} All extrapolation methods use only $15$--$20\%$ of the RLVR training cost, but \textsc{RELEX} is uniformly the strongest on MATH. Take Qwen2.5-Math-1.5B for example, \textsc{RELEX} beats Weight Extrapolation by $+1.2$ points, beats Logits Extrapolation by $+6.7$ points, beats ExPO by $+3.9$ points, and beats AlphaRL by $+4.3$ points. The Weight Extrapolation gap is the most informative: both methods rely on the empirical linearity of the trajectory (\S\ref{sec:preliminary}), but Weight Extrapolation fits a 2-point line directly on raw weight values, whereas \textsc{RELEX} first projects onto the rank-1 SVD subspace before extrapolating its scalar coefficient. This implies that the SVD step essentially acts as a low-pass filter that suppresses noisy residual directions that a raw 2-point fit absorbs as signal. AlphaRL also exploits rank-1 RL dynamics, but predicts the dominant update from checkpoint-level rank-1 components; \textsc{RELEX} instead performs a per-tensor trajectory SVD and fits the observed scalar coefficient over the full prefix, yielding stronger accuracy under the same compute budget in our setting. Moreover, the two-endpoint baselines substantially underperform our method, suggesting the advantage of \textsc{RELEX} in exploiting the full observed prefix.

\subsection{Ablation Studies and Analysis}
\label{sec:ablation}

\textsc{RELEX} has three design choices to justify: (1) operating in the SVD subspace rather than the raw weight space, (2) using a rank-1 projection (vs.\ higher rank), and (3) extrapolating with a linear function (vs.\ polynomial or neural). In Table~\ref{tab:ablation}, we ablate each on Qwen2.5-Math-1.5B with $\Tobs=75$, the same observation window used for the main comparison in Table~\ref{tab:main}.

\begin{table}[t]
\centering
\caption{Ablation studies on the design choices of \textsc{RELEX}, evaluating extrapolation performance on MATH across multiple target steps on Qwen2.5-Math-1.5B with $\Tobs{=}75$, and confirming the sufficiency of our default configuration (highlighted in \textbf{bold}).}
\label{tab:ablation}
\small
\begin{tabular}{llccccc}
\toprule
\textbf{Design choice} & \textbf{Variant} & \textbf{Step 100} & \textbf{Step 200} & \textbf{Step 300} & \textbf{Step 400} & \textbf{Step 500} \\
\midrule
Weight Space 
& \textbf{SVD (ours)} & \textbf{67.8} & \textbf{70.0} & \textbf{70.1} & \textbf{70.9} & \textbf{71.6} \\
& Raw & 67.4 & 68.5 & 69.8 & 70.4 & 70.7 \\
\midrule
Subspace rank & \textbf{Rank-1 (ours)} & \textbf{67.8} & \textbf{70.0} & \textbf{70.1} & \textbf{70.9} & \textbf{71.6} \\
              & Rank-5 & 67.0 & 68.4 & 69.9 & 69.8 & 70.6 \\
              & Rank-10 & 67.4 & 68.8 & 69.6 & 70.1 & 70.5 \\
\midrule
Function class & \textbf{Linear (ours)} & \textbf{67.8} & \textbf{70.0} & \textbf{70.1} & \textbf{70.9} & \textbf{71.6} \\
               & Polynomial & 66.9 & 17.8 & 0.2 & 0.2 & 0.1 \\
              & Neural Network & 67.2 & 69.5 & 70.5 & 70.5 & 72.1 \\
\bottomrule
\end{tabular}
\end{table}

\begin{figure}[t]
\centering
\includegraphics[width=\textwidth]{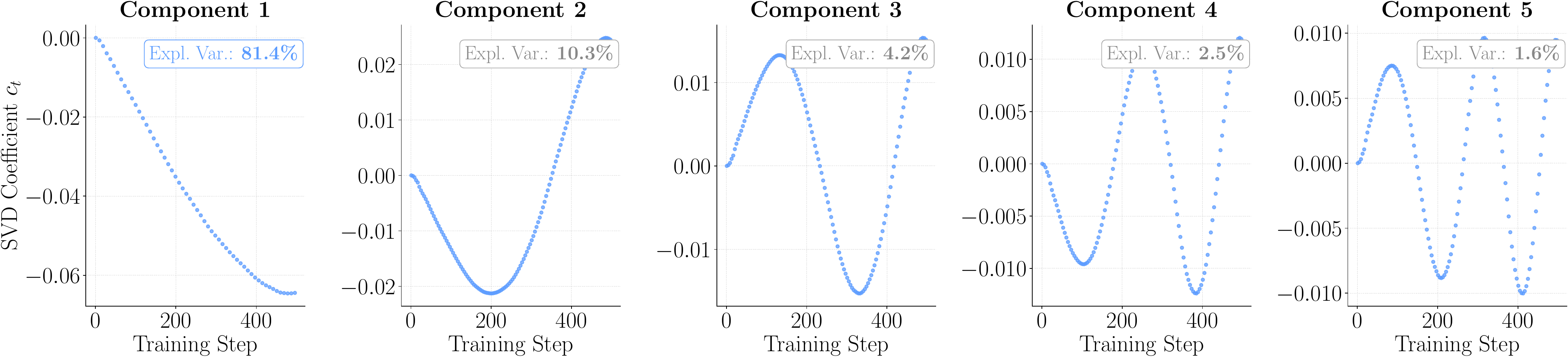}
\caption{Rank-5 SVD coefficient trajectories for a representative tensor (layer 14 \texttt{gate\_proj}, Qwen2.5-Math-1.5B). Annotation boxes show explained variance within the rank-5 subspace. Component 1 alone accounts for 81.4\% of the variance and evolves near-linearly over training, while components 2--5 together explain only 18.6\% and exhibit noisier dynamics.}
\label{fig:rank5_coeff}
\end{figure}

\paragraph{SVD projection acts as a spectral denoiser.}
\label{sec:regularization}
Table~\ref{tab:ablation} shows that when switching from the SVD space to the raw weight space, accuracy drops at every step. Moreover, the subspace rank ablation shows that adding components beyond the leading direction does not help, and Figure~\ref{fig:rank5_coeff} explains the mechanism: for a representative tensor, the leading rank-1 coefficient evolves smoothly across training steps and accounts for $81.4\%$ of the rank-5 subspace variance, while components 2--5 are lower-variance, noisier, and less monotonic. 
As a result, fitting in raw weight space reintroduces these noisy components, which extrapolation amplifies as drift.
In contrast, projecting onto the rank-1 subspace retains the smooth, monotone signals and discards the noisy ones.

\paragraph{Rank-1 is sufficient for extrapolation.}
\label{sec:rank1_linearity}

As shown in the subspace rank rows of Table~\ref{tab:ablation}, rank-5 and rank-10 fall behind rank-1 at every reported step. The added components do not compound a meaningful advantage. Figure~\ref{fig:rank5_coeff} clarifies why higher-rank fits do not help: the leading component is the only direction with a smooth, near-linear trajectory amenable to extrapolation, while components 2--5 behave too erratically for a linear fit to track reliably.
This echoes the preliminary observation in \S\ref{sec:preliminary} that rank-1 reconstruction already recovers full RLVR quality at every training step.
As a result, higher-rank components add modeling complexity but contribute little reliable extrapolation signal, which justifies \textsc{RELEX}'s rank-1 design: a single dominant scalar trajectory captures most of the structured dynamics needed for extrapolation.

\paragraph{Linear extrapolation outperforms more complicated functions.} 
We further compare three function families fit to the rank-1 coefficient trajectory: linear, polynomial (order 3), and a 3-layer neural network (Transformer)  trained to model the trajectory directly. The polynomial fit collapses catastrophically outside the observation window, and the neural network fit is competitive with linear but offers no consistent advantage at intermediate horizons (e.g., $69.5\%$ vs.\ $70.0\%$ at step 200) and incurs a much larger hyperparameter surface and per-step fitting cost.
As a result, we default \textsc{RELEX} to linear extrapolation due to its simplicity---it admits a closed-form least-squares solution with no learnable parameters, and the empirical observation of linearity in the rank-1 coefficient (\S\ref{sec:preliminary}).

\paragraph{\textsc{RELEX} extrapolates stably far beyond the observed window.}
\label{sec:analysis}
\label{sec:hyperparam_tobs}
\begin{table}[t]
\centering
\small
\caption{Observation-window sweep ($\Tobs \in \{50, 75, 100, 125\}$) and long-horizon stability of \textsc{RELEX}. Entries report MATH accuracy (\%) at extrapolated steps, with Base and RLVR accuracy for reference.
``---'' indicates no extrapolated checkpoint at that step.
The best step in each row is marked in \textbf{bold}.
}
\label{tab:cutoff_combined}
\begin{tabular}{lcc|cccccccc}
\toprule
\textbf{Model} & \textbf{Base} & \textbf{RLVR} & $\boldsymbol{\Tobs}$ & \textbf{100} & \textbf{200} & \textbf{300} & \textbf{400} & \textbf{500} & \textbf{750} & \textbf{1000} \\
\midrule
\multirow{4}{*}{Qwen2.5-Math-1.5B} & \multirow{4}{*}{48.2} & \multirow{4}{*}{71.5}
 &  50 & 67.7 & 68.5 & 69.2 & 70.2 & \textbf{70.8} & 70.8 & 66.3 \\
 &                       &                          & 75 & 67.8 & 70.0 & 70.1 & 70.9 & \textbf{71.6} & 65.7 & 68.4 \\
 &                       &                          & 100 & ---  & 68.6 & 69.5 & 70.2 & 70.4 & \textbf{71.3} & 71.2 \\
 &                       &                          & 125 & ---  & 69.1 & 69.9 & 70.5 & 70.9 & \textbf{71.7} & 71.6 \\
\midrule
\multirow{4}{*}{Qwen3-4B-Base} & \multirow{4}{*}{64.0} & \multirow{4}{*}{85.5}
 &  50 & 79.8 & 81.7 & 82.4 & 83.8 & \textbf{84.1} & 82.5 & 78.3 \\
 &                       &                          & 75 & 79.6 & 81.7 & 82.8 & 84.2 & \textbf{85.6} & 82.7 & 73.6 \\
 &                       &                          & 100 & ---  & 83.7 & \textbf{85.2} & 85.3 & 85.0 & 79.0 & 61.3 \\
 &                       &                          & 125 & ---  & 83.7 & 84.6 & \textbf{84.9} & 83.7 & 73.8 & 50.9 \\
\midrule
\multirow{4}{*}{Qwen3-8B-Base} & \multirow{4}{*}{73.9} & \multirow{4}{*}{88.5}
 &  50 & 84.6 & 84.9 & 85.9 & \textbf{86.2} & 84.7 & 57.4 & 22.7 \\
 &                       &                          & 75 & 83.7 & 85.9 & 86.2 & 86.3 & \textbf{87.0} & 78.3 & 50.4 \\
 &                       &                          & 100 & ---  & 85.6 & 85.9 & 86.4 & \textbf{87.7} & 87.4 & 82.4 \\
 &                       &                          & 125 & ---  & 86.0 & 85.9 & 86.6 & 87.5 & \textbf{88.0} & 85.6 \\
\bottomrule
\end{tabular}
\end{table}

Table~\ref{tab:cutoff_combined} sweeps the observation window $\Tobs \in \{50, 75, 100, 125\}$ jointly with the target extrapolation step across all three models. Under a well-chosen $\Tobs$, \textsc{RELEX} remains close to peak accuracy as far out as step 1000, which is roughly $8\times$ the observation window and twice the original 500-step RLVR horizon. For example, Qwen2.5-Math-1.5B with $\Tobs{=}125$ peaks at step 750 and stays at $71.6$\% at step 1000, exceeding the $71.5$\% RLVR step-500 reference. Likewise Qwen3-8B-Base with $\Tobs{=}125$ peaks at step 750 and remains at $85.6$\% at step 1000. 
Note that the choice of $\Tobs$ is consequential, and the right value differs by model: smaller $\Tobs{=}75$ destabilizes long-horizon extrapolation on Qwen2.5-Math-1.5B (drops to $65.7$\% at step 750) and Qwen3-8B-Base (drops to $50.4$\% at step 1000), whereas larger windows track the trajectory cleanly. Qwen3-4B is the harder case---no $\Tobs$ in this sweep sustains accuracy beyond step 750 ($\Tobs{=}75$ still scores $82.7$\% at step 750 but falls to $73.6$\% at step 1000, while larger windows degrade much earlier, dropping to $61.3$\% and $50.9$\% at step 1000), suggesting that long-horizon extrapolation stability requires a matched observation window for each model.

%% file: tex/5-related.tex
\section{Related Work}
\label{sec:related}

\paragraph{Structure of RLVR training dynamics.} Some recent works study the geometry and optimization dynamics of RLVR training. \citet{zhu2025rlvr} analyze RLVR through the lens of principal components, showing that RL learns off the principals, while \citet{huang2026beyond} argue that the \emph{direction} of RLVR updates matters more than their magnitude. \citet{mukherjee2025subnetworks} find that RL finetunes only a few portions of parameters and \citet{ye2026implicit} further study rank-1 components in RLVR and connect low-rank dynamics to implicit reward overfitting and singular-spectrum changes. \citet{shenfeld2026rls} provide theoretical justification via RL's Razor: on-policy RL is implicitly biased toward KL-minimal solutions, which may explain why RLVR updates remain low-rank. \citet{huang2026implicit} analyze RLVR learning dynamics from a complementary theoretical perspective, showing how mixed-difficulty data induces an implicit curriculum. On the extrapolation side, \citet{zheng2025expo} amplify a two-endpoint weight displacement to accelerate training. Most closely related, \citet{wang2026linearity} observe that both weights and logits evolve linearly during RLVR, and propose Weight Extrapolation and Logits Extrapolation to reduce training cost. Our work shares the core linearity observation but differs in two key respects. First, \citet{wang2026linearity} extrapolate raw weight values using only two checkpoints (base and one intermediate), which is sensitive to noise in a single delta and treats each weight independently. \textsc{RELEX} instead fits ordinary least squares over all observed steps and operates in the rank-1 SVD subspace, which (i) makes the slope estimate more robust to per-step noise and (ii) discards high-frequency weight components that do not contribute to task performance, acting as a spectral denoiser. Second, \citet{wang2026linearity} observe linearity at the level of individual raw weights (R$^2 > 0.7$ for 80\% of weights), while we show that the rank-1 SVD coefficient achieves R$^2 > 0.98$ across most tensors, showing a higher signal of regularity that directly motivates the rank-1 projection in \textsc{RELEX}.

\paragraph{Low-rank structure and weight-space modeling.} The low-rank nature of weight updates has been observed in supervised fine-tuning~\citep{li2018measuring,aghajanyan2021intrinsic} and exploited by LoRA~\citep{hu2022lora}. In classical deep RL, \citet{tang2024the} show that policy learning concentrates along a small number of major parameter directions. Closest in spirit, \citet{cai2026on} identify rank-1 dominance and rank-1 linear dynamics in RL-induced updates and propose AlphaRL to accelerate training by predicting the dominant update from early checkpoints. Crucially, AlphaRL computes a separate SVD of $\dt_t$ at each observed checkpoint and uses these per-checkpoint decompositions, so the rank-1 basis is re-derived at every step and may rotate across the trajectory. \textsc{RELEX} instead performs a single per-tensor \emph{trajectory} SVD over the entire observation window, yielding one \emph{shared} rank-1 basis $\mathbf{v}_1$ along which the scalar coefficient $c_t$ evolves near-linearly. \citet{chen2026low} argue that the rank-1 subspace need not evolve linearly and propose NExt, which trains a predictor over low-rank LoRA trajectories for nonlinear extrapolation. In contrast to these learned or checkpoint-level predictors, \textsc{RELEX} uses a trajectory-level SVD and a closed-form linear fit of the rank-1 coefficient, requiring no predictor training or extra modules. In the broader context of weight-space modeling, \citet{li2026weightflow} model dynamics directly in weight space via graph controlled differential equations, while \citet{zeng2025generative} caution that generative models of neural network weights tend to memorize rather than generalize, underscoring the difficulty of this domain. 

\paragraph{Model merging and scaling laws.} Task arithmetic~\citep{ilharco2023editing} and model soups~\citep{wortsman2022model} exploit linear structure in weight space for model combination, while LoraHub~\citep{huang2024lorahub} composes task-specific LoRA modules for cross-task generalization. These methods operate on static endpoints (independently trained models or adapters); our work extends the paradigm from interpolation between endpoints to \emph{extrapolation} along a single model's evolving training trajectory. Scaling laws~\citep{kaplan2020scaling,hoffmann2022training} predict aggregate loss from compute, while our work predicts full model parameters from early dynamics.

%% file: tex/6-discussion.tex
\section{Discussion}
\label{sec:discussion}

\paragraph{Observation-window sensitivity.} The cutoff sweeps show that the best observation window is model-dependent. Qwen2.5-Math-1.5B benefits from longer prefixes, while Qwen3-family models are more sensitive to which checkpoints are included in the SVD fit: small or intermediate windows can outperform longer ones, and early-window extrapolations may become unstable at long horizons. This suggests that the dominant rank-1 direction estimated from a short prefix is not equally stationary across model families. A practical next step is to monitor subspace drift or singular-value gaps online and select $\Tobs$ adaptively.

\paragraph{Subspace selection is critical.} As discussed in the ablation study, the subspace matters for \textsc{RELEX}, and the right choice depends on model. On Qwen3-4B-Base, early-subspace (\eg, $\Tobs = 75$) remain competitive with RLVR, but larger cutoffs degrade. The pattern reverses on Qwen3-8B-Base, where intermediate-to-large windows ($\Tobs \in \{100, 125\}$) recover the most RLVR-level accuracy and small windows underperform. The practical implication is twofold: SVD projection serves as a near-zero-cost post-processing step for completed RLVR runs, but the choice of subspace (early vs.\ full-trajectory, rank-1 vs.\ rank-$r$, small vs.\ large $\Tobs$) must be validated empirically per model family.

\paragraph{Limitations.} In this paper, we mainly study RLVR with GRPO on mathematical reasoning across three Qwen-family models. Whether similar low-rank structure holds for other RL algorithms (\eg, PPO), other task forms (\eg, code generation), or other model families (\eg, Llama) remains open. The rank-1 design choice, while sufficient for the studied models with appropriate cutoffs, may not generalize universally, motivating adaptive rank selection as an important future direction.

%% file: tex/7-conclusion.tex
\section{Conclusion}
\label{sec:conclusion}

In this paper, we show that RLVR weight updates follow low-rank, predictable trajectories: parameter deltas concentrate in a rank-1 subspace per tensor, and the scalar coefficient evolves near-linearly with training step. This geometric regularity enables \textsc{RELEX}, a simple parameter-free method that extrapolates future checkpoints via rank-1 SVD followed by linear regression, with no learned model required. With $15$--$20\%$ of training observed, \textsc{RELEX} matches RLVR on in-domain MATH for Qwen2.5-Math-1.5B, Qwen3-4B-Base and Qwen3-8B-Base, while matching or improving out-of-distribution accuracy across the board. Analysis shows that SVD projection itself also acts as a beneficial spectral regularizer that can improve upon the original checkpoints. The opposite observation-window preferences on Qwen3-4B-Base (small $\Tobs$) and Qwen3-8B-Base (large $\Tobs$) suggest that rank-1 stationarity is model-dependent, motivating adaptive subspace selection as a key next step. More broadly, our findings suggest that RL training, despite its stochastic and non-convex nature, carves surprisingly simple paths in low-rank parameter space, exhibiting predictable patterns that can be exploited for efficient checkpoint extrapolation and deeper understanding of training dynamics.

%% file: tex/appendix.tex
\section{Implementation Details}
\label{app:implementation}

\paragraph{RLVR training details.} We train models with the verl framework~\citep{sheng2025hybridflow} using GRPO on the MATH training split. Unless otherwise stated, training uses AdamW with learning rate $10^{-6}$, KL coefficient 0.001, group size $G=8$ rollouts per prompt, and mini-batch size 256. All runs across three models (\ie, Qwen2.5-Math-1.5B, Qwen3-4B-Base, Qwen3-8B-Base) are trained for 500 optimization steps on 8xH200 GPUs.

\paragraph{Inference details.} For in-domain MATH evaluation, Qwen2.5-Math-1.5B uses greedy decoding with a 4K-token budget, while Qwen3-family models use sampling decoding with a 16K-token budget. For OOD benchmarks, we use avg@8 sampling with temperature of 0.7. All methods compared within the same model block use the same inference settings.

\paragraph{SVD computation.} For each weight tensor $W^{(\ell)}$, we form deltas $\Delta_t^{(\ell)} = W_t^{(\ell)} - W_0^{(\ell)}$ and stack flattened deltas into a trajectory matrix $\mathbf{M}^{(\ell)} \in \mathbb{R}^{T \times d_\ell}$. Since $T \ll d_\ell$, we compute the compact SVD through the $T \times T$ Gram matrix $\mathbf{G}^{(\ell)} = \mathbf{M}^{(\ell)}\mathbf{M}^{(\ell)\top}$, then recover the right singular directions as needed. We cache per-tensor singular directions and coefficient trajectories to avoid recomputing SVDs across cutoff sweeps.

\paragraph{Comparison with extrapolation baselines.} Weight Extrapolation and Logits Extrapolation follow the two-endpoint extrapolation setup of \citet{wang2026linearity}: they use two checkpoints $\theta_{t_0}$ and $\theta_{\Tobs}$, linearly extrapolating either raw weights or output logits. ExPO~\citep{zheng2025expo} amplifies the displacement from an initial checkpoint to a partially trained checkpoint. AlphaRL~\citep{cai2026on} estimates a rank-1 SVD of the delta independently at each observed checkpoint and uses these per-checkpoint decompositions to predict the dominant update vector at the target step. As a result, the rank-1 basis is re-derived per step and may rotate across the trajectory. In contrast, \textsc{RELEX} uses the full observed checkpoint trajectory to estimate both a per-tensor rank-1 subspace and its temporal coefficient dynamics. A detailed comparison of these methods is provided in Table~\ref{tab:method_compare}.

\begin{table}[t]
\centering
\small
\caption{Comparison of extrapolation methods. ExPO, Weight Extrapolation, and Logits Extrapolation rely on two endpoint checkpoints, capturing only a static displacement. AlphaRL computes a separate rank-1 SVD at each observed checkpoint and uses PLS regression to predict a single dominant update vector. In contrast, \textsc{RELEX} performs a single per-tensor SVD over the full checkpoint trajectory (one shared rank-1 basis) and fits the scalar coefficient with closed-form linear regression---capturing both the dominant subspace and its temporal dynamics, which together enable structured denoising and robust extrapolation.}
\label{tab:method_compare}
\begingroup
\renewcommand{\arraystretch}{2}
\resizebox{\textwidth}{!}{%
\begin{tabular}{lccc}
\toprule
\textbf{Method} & \textbf{Space} & \textbf{Extrapolation format} & \textbf{Prior information} \\
\midrule
ExPO~\citep{zheng2025expo} & Raw weights & $\hat\theta=\theta_{\Tobs}+\alpha(\theta_{\Tobs}-\theta_0)$ & Two endpoints: $\{\theta_0, \theta_{\Tobs}\}$ \\
Weight Extrap.~\citep{wang2026linearity} & Raw weights & $\hat\theta_T=\theta_{t_0}+\frac{T-t_0}{\Tobs-t_0}(\theta_{\Tobs}-\theta_{t_0})$ & Two endpoints: $\{\theta_{t_0}, \theta_{\Tobs}\}$ \\
Logits Extrap.~\citep{wang2026linearity} & Output logits & $\hat z_T=z_{t_0}+\frac{T-t_0}{\Tobs-t_0}(z_{\Tobs}-z_{t_0})$ & Two endpoints: $\{\theta_{t_0}, \theta_{\Tobs}\}$ \\
AlphaRL~\citep{cai2026on} & \shortstack{Rank-1 vector \\ (per-checkpoint)} & $\hat\theta_T=\theta_{0}+\mathbf{u}\,\mathbf{v}_1^{(\Tobs)},\;\mathbf{u}=\mathrm{PLS}^{-1}(y{=}1)$ & \shortstack{$\{\theta_t\}_{t=0}^{\Tobs}$, per-checkpoint\\SVDs (independent)} \\
{\bf \textsc{RELEX} (ours)} & \shortstack{Rank-1 coefficient\\ (per-trajectory)} & $\hat\theta_T=\theta_0+\hat c_T \mathbf{v}_1,\;\hat c_T=aT+b$ & \shortstack{$\{\theta_t\}_{t=0}^{\Tobs}$, per-trajectory\\SVD (shared basis $\mathbf{v}_1$)} \\
\bottomrule
\end{tabular}}
\endgroup
\end{table}

\section{Weight-Space Alignment Analysis}
\label{app:direction_alignment}

\paragraph{Reconstruction tracks the RLVR trajectory in raw weight space while extrapolation drifts in both direction and magnitude as the horizon grows.}
We analyze the alignment between extrapolated checkpoints and actual RLVR checkpoints in raw weight space under two settings: \emph{Reconstruction} ($\Tobs=500$, Algorithm~\ref{alg:decompose}), in which SVD is fit on the entire trajectory and $\hat{W}(t)$ is the rank-1 projection of $W_{\mathrm{RLVR}}(t)$, and \emph{Extrapolation} ($\Tobs=75$, Algorithm~\ref{alg:rlex}), in which SVD is fit on the first 75 steps and the coefficient $c_1(t)$ is linearly extrapolated to predict $\hat{W}(t)$ for $t>75$ without ever seeing $W_{\mathrm{RLVR}}(t)$. For each checkpoint step $t \in \{100, 200, 300, 400, 500\}$, we compute $\Delta_{\mathrm{RLVR}}(t) = W_{\mathrm{RLVR}}(t) - W_0$ and $\hat{\Delta}(t) = \hat{W}(t) - W_0$, and report the mean per-tensor cosine similarity and the norm ratio $\|\hat{\Delta}(t)\| / \|\Delta_{\mathrm{RLVR}}(t)\|$.

\begin{figure}[t]
\centering
\begin{subfigure}[t]{0.45\textwidth}
    \centering
    \includegraphics[width=\textwidth]{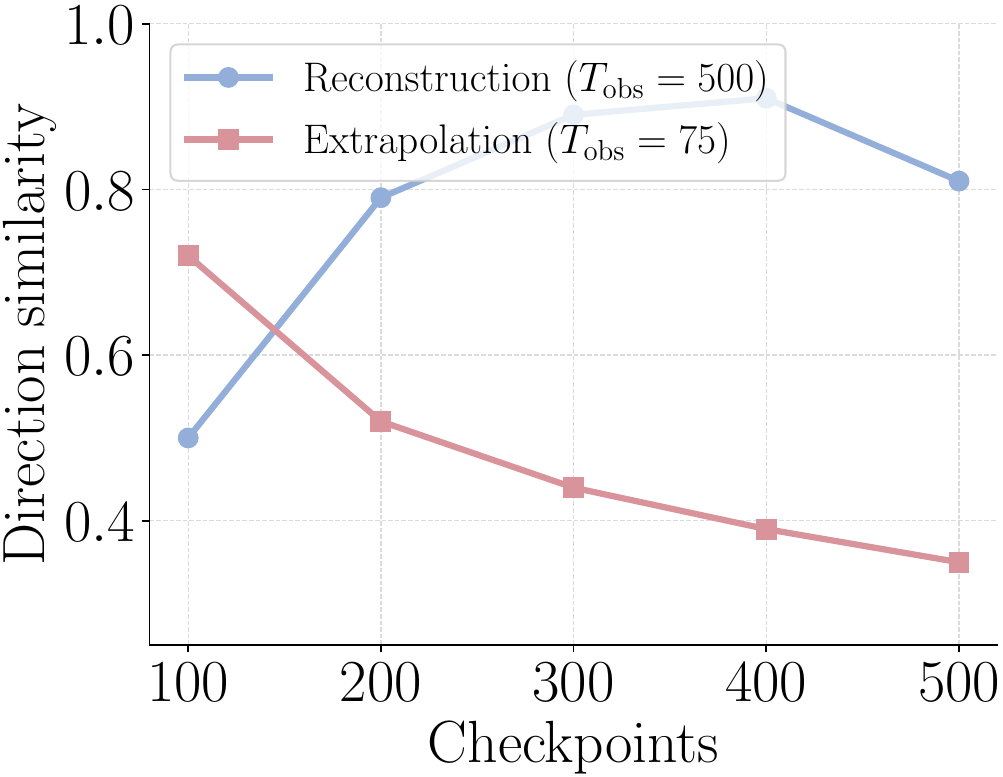}
    \caption{Direction similarity.}
    \label{fig:direction_alignment_dir}
\end{subfigure}
\hfill
\begin{subfigure}[t]{0.45\textwidth}
    \centering
    \includegraphics[width=\textwidth]{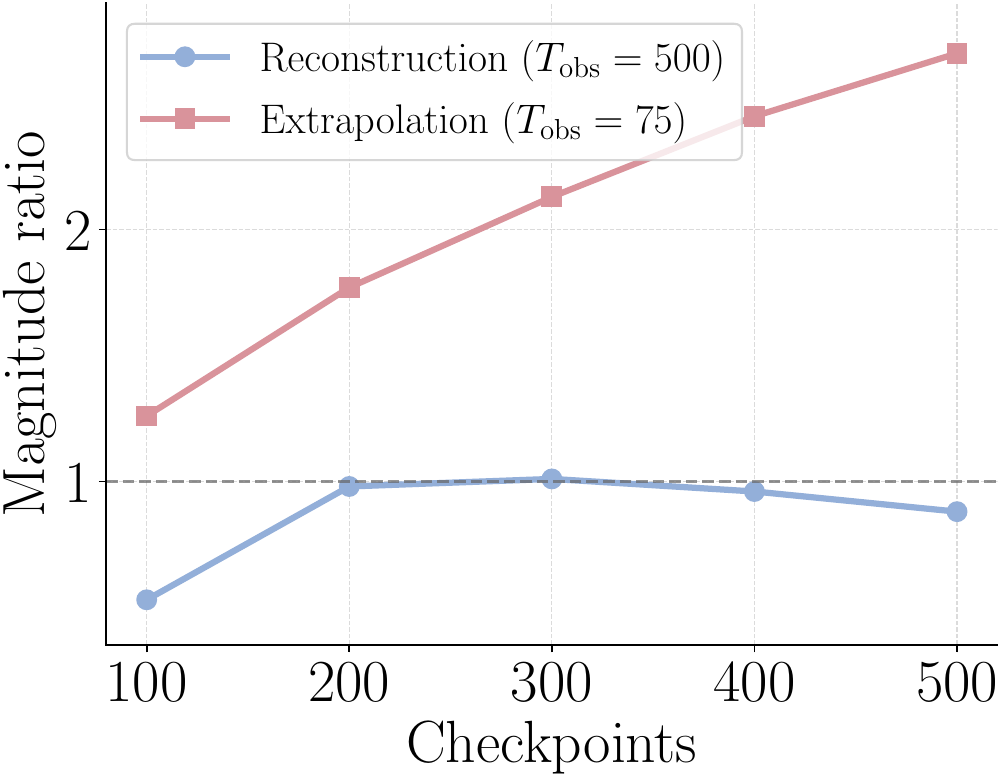}
    \caption{Magnitude ratio.}
    \label{fig:direction_alignment_mag}
\end{subfigure}
\caption{\textbf{Weight-space alignment against the true RLVR trajectory on Qwen2.5-Math-1.5B.} Reconstruction ($\Tobs=500$, Algorithm~\ref{alg:decompose}) is a rank-1 projection of the actual delta within the observed window; extrapolation ($\Tobs=75$, Algorithm~\ref{alg:rlex}) fits the first 75 steps and predicts future checkpoints without seeing $W_{\mathrm{RLVR}}(t)$. \textbf{(a)} mean per-tensor direction similarity (cosine to $\Delta_{\mathrm{RLVR}}(t)$); \textbf{(b)} magnitude ratio $\|\hat{\Delta}\|/\|\Delta_{\mathrm{RLVR}}\|$.}
\label{fig:direction_alignment}
\vspace{-0.75em}
\end{figure}

Figure~\ref{fig:direction_alignment} shows two regimes. Under the reconstruction setting, the rank-1 subspace recovers the dominant RLVR direction with high directional similarity ($0.50 \to 0.91$, peaking near step 400 when the trajectory is best aligned with the dominant singular vector $\mathbf{v}_1$) and a magnitude ratio close to $1.0$, confirming that the rank-1 projection faithfully captures the trajectory within the observed prefix. The lower direction similarity at step 100 reflects that early RLVR updates are noisier and less concentrated along $\mathbf{v}_1$, which is estimated from the full 500-step trajectory.
Under the extrapolation setting, direction similarity decays and magnitude over-extrapolation grows monotonically with horizon (direction sim.\ $0.72 \to 0.35$, magnitude ratio $1.26 \to 2.70$ at $t=500$), reflecting a genuine drift from the true RLVR trajectory in raw weight space. 

%% file: neurips_2026.bbl
\begin{thebibliography}{34}
\providecommand{\natexlab}[1]{#1}
\providecommand{\url}[1]{\texttt{#1}}
\expandafter\ifx\csname urlstyle\endcsname\relax
  \providecommand{\doi}[1]{doi: #1}\else
  \providecommand{\doi}{doi: \begingroup \urlstyle{rm}\Url}\fi

\bibitem[Aghajanyan et~al.(2021)Aghajanyan, Gupta, and Zettlemoyer]{aghajanyan2021intrinsic}
Armen Aghajanyan, Sonal Gupta, and Luke Zettlemoyer.
\newblock Intrinsic dimensionality explains the effectiveness of language model fine-tuning.
\newblock In \emph{Proceedings of the 59th annual meeting of the association for computational linguistics and the 11th international joint conference on natural language processing (volume 1: long papers)}, pages 7319--7328, 2021.

\bibitem[Cai et~al.(2026)Cai, Cao, Xu, Yao, Huang, Tan, Zhang, Liu, and Fang]{cai2026on}
Yuchen Cai, Ding Cao, Xin Xu, Zijun Yao, Yuqing Huang, Zhenyu Tan, Benyi Zhang, Guiquan Liu, and Junfeng Fang.
\newblock On predictability of reinforcement learning dynamics for large language models.
\newblock In \emph{The Fourteenth International Conference on Learning Representations}, 2026.

\bibitem[Chen et~al.(2026)Chen, Qian, Zhao, and Wen]{chen2026low}
Zhipeng Chen, Tao Qian, Wayne~Xin Zhao, and Ji-Rong Wen.
\newblock Low-rank optimization trajectories modeling for {LLM} {RLVR} acceleration.
\newblock \emph{arXiv preprint arXiv:2604.11446}, 2026.

\bibitem[Dekoninck et~al.(2026)Dekoninck, Jovanović, Gehrunger, Rögnvaldsson, Petrov, Sun, and Vechev]{dekoninck2026matharena}
Jasper Dekoninck, Nikola Jovanović, Tim Gehrunger, Kári Rögnvaldsson, Ivo Petrov, Chenhao Sun, and Martin Vechev.
\newblock Beyond benchmarks: {MathArena} as an evaluation platform for mathematics with llms.
\newblock \emph{arXiv preprint arXiv:2605.00674}, 2026.

\bibitem[Guo et~al.(2025)Guo, Yang, Zhang, Song, Wang, Zhu, Xu, Zhang, Ma, Bi, et~al.]{guo2025deepseek}
Daya Guo, Dejian Yang, Haowei Zhang, Junxiao Song, Peiyi Wang, Qihao Zhu, Runxin Xu, Ruoyu Zhang, Shirong Ma, Xiao Bi, et~al.
\newblock {DeepSeek-R1}: Incentivizing reasoning capability in {LLMs} via reinforcement learning.
\newblock \emph{arXiv preprint arXiv:2501.12948}, 2025.

\bibitem[He et~al.(2024)He, Luo, Bai, Hu, Thai, Shen, Hu, Han, Huang, Zhang, et~al.]{he2024olympiadbench}
Chaoqun He, Renjie Luo, Yuzhuo Bai, Shengding Hu, Zhen Thai, Junhao Shen, Jinyi Hu, Xu~Han, Yujie Huang, Yuxiang Zhang, et~al.
\newblock {OlympiadBench}: A challenging benchmark for promoting {AGI} with olympiad-level bilingual multimodal scientific problems.
\newblock In \emph{Proceedings of the 62nd Annual Meeting of the Association for Computational Linguistics (Volume 1: Long Papers)}, pages 3828--3850, 2024.

\bibitem[Hendrycks et~al.(2021)Hendrycks, Burns, Kadavath, Arora, Basart, Tang, Song, and Steinhardt]{hendrycks2021math}
Dan Hendrycks, Collin Burns, Saurav Kadavath, Akul Arora, Steven Basart, Eric Tang, Dawn Song, and Jacob Steinhardt.
\newblock Measuring mathematical problem solving with the {MATH} dataset.
\newblock In \emph{Thirty-fifth Conference on Neural Information Processing Systems Datasets and Benchmarks Track (Round 2)}, 2021.

\bibitem[Hoffmann et~al.(2022)Hoffmann, Borgeaud, Mensch, Buchatskaya, Cai, Rutherford, de~las Casas, Hendricks, Welbl, Clark, Hennigan, Noland, Millican, van~den Driessche, Damoc, Guy, Osindero, Simonyan, Elsen, Vinyals, Rae, and Sifre]{hoffmann2022training}
Jordan Hoffmann, Sebastian Borgeaud, Arthur Mensch, Elena Buchatskaya, Trevor Cai, Eliza Rutherford, Diego de~las Casas, Lisa~Anne Hendricks, Johannes Welbl, Aidan Clark, Tom Hennigan, Eric Noland, Katherine Millican, George van~den Driessche, Bogdan Damoc, Aurelia Guy, Simon Osindero, Karen Simonyan, Erich Elsen, Oriol Vinyals, Jack~William Rae, and Laurent Sifre.
\newblock Training compute-optimal large language models.
\newblock In \emph{Advances in Neural Information Processing Systems}, 2022.

\bibitem[Hu et~al.(2022)Hu, Shen, Wallis, Allen-Zhu, Li, Wang, Wang, and Chen]{hu2022lora}
Edward~J Hu, Yelong Shen, Phillip Wallis, Zeyuan Allen-Zhu, Yuanzhi Li, Shean Wang, Lu~Wang, and Weizhu Chen.
\newblock Lo{RA}: Low-rank adaptation of large language models.
\newblock In \emph{International Conference on Learning Representations}, 2022.

\bibitem[Huang et~al.(2024)Huang, Liu, Lin, Pang, Du, and Lin]{huang2024lorahub}
Chengsong Huang, Qian Liu, Bill~Yuchen Lin, Tianyu Pang, Chao Du, and Min Lin.
\newblock {LoraHub}: Efficient cross-task generalization via dynamic {LoRA} composition.
\newblock In \emph{First Conference on Language Modeling}, 2024.

\bibitem[Huang et~al.(2026{\natexlab{a}})Huang, Meng, Wu, Lu, Ma, Chen, Wang, Ding, Wu, Wang, He, Wang, and Zhou]{huang2026beyond}
Kexin Huang, Haoming Meng, Junkang Wu, Jinda Lu, Chiyu Ma, Ziqian Chen, Xue Wang, Bolin Ding, Jiancan Wu, Xiang Wang, Xiangnan He, Guoyin Wang, and Jingren Zhou.
\newblock Beyond magnitude: Leveraging direction of {RLVR} updates for {LLM} reasoning.
\newblock In \emph{The Fourteenth International Conference on Learning Representations}, 2026{\natexlab{a}}.

\bibitem[Huang et~al.(2026{\natexlab{b}})Huang, Wen, Chi, Wei, Singh, Liang, and Chen]{huang2026implicit}
Yu~Huang, Zixin Wen, Yuejie Chi, Yuting Wei, Aarti Singh, Yingbin Liang, and Yuxin Chen.
\newblock The implicit curriculum: Learning dynamics in {RL} with verifiable rewards.
\newblock \emph{arXiv preprint arXiv:2602.14872}, 2026{\natexlab{b}}.

\bibitem[Ilharco et~al.(2023)Ilharco, Ribeiro, Wortsman, Schmidt, Hajishirzi, and Farhadi]{ilharco2023editing}
Gabriel Ilharco, Marco~Tulio Ribeiro, Mitchell Wortsman, Ludwig Schmidt, Hannaneh Hajishirzi, and Ali Farhadi.
\newblock Editing models with task arithmetic.
\newblock In \emph{The Eleventh International Conference on Learning Representations}, 2023.

\bibitem[Kaplan et~al.(2020)Kaplan, McCandlish, Henighan, Brown, Chess, Child, Gray, Radford, Wu, and Amodei]{kaplan2020scaling}
Jared Kaplan, Sam McCandlish, Tom Henighan, Tom~B Brown, Benjamin Chess, Rewon Child, Scott Gray, Alec Radford, Jeffrey Wu, and Dario Amodei.
\newblock Scaling laws for neural language models.
\newblock \emph{arXiv preprint arXiv:2001.08361}, 2020.

\bibitem[Lambert et~al.(2025)Lambert, Morrison, Pyatkin, Huang, Ivison, Brahman, Miranda, Liu, Dziri, Lyu, Gu, Malik, Graf, Hwang, Yang, Bras, Tafjord, Wilhelm, Soldaini, Smith, Wang, Dasigi, and Hajishirzi]{lambert2025tulu}
Nathan Lambert, Jacob Morrison, Valentina Pyatkin, Shengyi Huang, Hamish Ivison, Faeze Brahman, Lester James~Validad Miranda, Alisa Liu, Nouha Dziri, Xinxi Lyu, Yuling Gu, Saumya Malik, Victoria Graf, Jena~D. Hwang, Jiangjiang Yang, Ronan~Le Bras, Oyvind Tafjord, Christopher Wilhelm, Luca Soldaini, Noah~A. Smith, Yizhong Wang, Pradeep Dasigi, and Hannaneh Hajishirzi.
\newblock Tulu 3: Pushing frontiers in open language model post-training.
\newblock In \emph{Second Conference on Language Modeling}, 2025.

\bibitem[Li et~al.(2018)Li, Farkhoor, Liu, and Yosinski]{li2018measuring}
Chunyuan Li, Heerad Farkhoor, Rosanne Liu, and Jason Yosinski.
\newblock Measuring the intrinsic dimension of objective landscapes.
\newblock In \emph{International Conference on Learning Representations}, 2018.

\bibitem[Li et~al.(2026)Li, Liu, Wang, Liao, and Li]{li2026weightflow}
Ruikun Li, Jiazhen Liu, Huandong Wang, Qingmin Liao, and Yong Li.
\newblock {WeightFlow}: Learning stochastic dynamics via evolving weight of neural network.
\newblock In \emph{Proceedings of the AAAI Conference on Artificial Intelligence}, volume~40, pages 641--649, 2026.

\bibitem[Liu et~al.(2025)Liu, Diao, Lu, Hu, Dong, Choi, Kautz, and Dong]{liu2025prorl}
Mingjie Liu, Shizhe Diao, Ximing Lu, Jian Hu, Xin Dong, Yejin Choi, Jan Kautz, and Yi~Dong.
\newblock Pro{RL}: Prolonged reinforcement learning expands reasoning boundaries in large language models.
\newblock In \emph{The Thirty-ninth Annual Conference on Neural Information Processing Systems}, 2025.

\bibitem[Mukherjee et~al.(2025)Mukherjee, Yuan, Hakkani-T{\"u}r, and Peng]{mukherjee2025subnetworks}
Sagnik Mukherjee, Lifan Yuan, Dilek Hakkani-T{\"u}r, and Hao Peng.
\newblock Reinforcement learning finetunes small subnetworks in large language models.
\newblock In \emph{The Thirty-ninth Annual Conference on Neural Information Processing Systems}, 2025.

\bibitem[Olmo et~al.(2025)Olmo, Ettinger, Bertsch, Kuehl, Graham, Heineman, Groeneveld, Brahman, Timbers, Ivison, et~al.]{olmo2025olmo}
Team Olmo, Allyson Ettinger, Amanda Bertsch, Bailey Kuehl, David Graham, David Heineman, Dirk Groeneveld, Faeze Brahman, Finbarr Timbers, Hamish Ivison, et~al.
\newblock Olmo 3.
\newblock \emph{arXiv preprint arXiv:2512.13961}, 2025.

\bibitem[Shao et~al.(2024)Shao, Wang, Zhu, Xu, Song, Bi, Zhang, Zhang, Li, Wu, et~al.]{shao2024deepseekmath}
Zhihong Shao, Peiyi Wang, Qihao Zhu, Runxin Xu, Junxiao Song, Xiao Bi, Haowei Zhang, Mingchuan Zhang, YK~Li, Yang Wu, et~al.
\newblock {DeepSeekMath}: Pushing the limits of mathematical reasoning in open language models.
\newblock \emph{arXiv preprint arXiv:2402.03300}, 2024.

\bibitem[Shenfeld et~al.(2026)Shenfeld, Pari, and Agrawal]{shenfeld2026rls}
Idan Shenfeld, Jyothish Pari, and Pulkit Agrawal.
\newblock {RL}'s razor: Why online reinforcement learning forgets less.
\newblock In \emph{The Fourteenth International Conference on Learning Representations}, 2026.

\bibitem[Sheng et~al.(2025)Sheng, Zhang, Ye, Wu, Zhang, Zhang, Peng, Lin, and Wu]{sheng2025hybridflow}
Guangming Sheng, Chi Zhang, Zilingfeng Ye, Xibin Wu, Wang Zhang, Ru~Zhang, Yanghua Peng, Haibin Lin, and Chuan Wu.
\newblock {HybridFlow}: A flexible and efficient {RLHF} framework.
\newblock In \emph{Proceedings of the Twentieth European Conference on Computer Systems}, pages 1279--1297, 2025.

\bibitem[Tang et~al.(2024)Tang, Zhang, Chen, and Hao]{tang2024the}
Hongyao Tang, Min Zhang, Chen Chen, and Jianye Hao.
\newblock The ladder in chaos: Improving policy learning by harnessing the parameter evolving path in a low-dimensional space.
\newblock In \emph{The Thirty-eighth Annual Conference on Neural Information Processing Systems}, 2024.

\bibitem[Wang et~al.(2026)Wang, Wu, Jin, Xu, Chen, and Miao]{wang2026linearity}
Tianle Wang, Zhongyuan Wu, Shenghao Jin, Hao Xu, Wei Chen, and Ning Miao.
\newblock Not all steps are informative: On the linearity of {LLMs}' {RLVR} training.
\newblock \emph{arXiv preprint arXiv:2601.04537}, 2026.

\bibitem[Wortsman et~al.(2022)Wortsman, Ilharco, Gadre, Roelofs, Gontijo-Lopes, Morcos, Namkoong, Farhadi, Carmon, Kornblith, et~al.]{wortsman2022model}
Mitchell Wortsman, Gabriel Ilharco, Samir~Ya Gadre, Rebecca Roelofs, Raphael Gontijo-Lopes, Ari~S Morcos, Hongseok Namkoong, Ali Farhadi, Yair Carmon, Simon Kornblith, et~al.
\newblock Model soups: averaging weights of multiple fine-tuned models improves accuracy without increasing inference time.
\newblock In \emph{International conference on machine learning}, pages 23965--23998, 2022.

\bibitem[Yang et~al.(2024)Yang, Zhang, Hui, Gao, Yu, Li, Liu, Tu, Zhou, Lin, et~al.]{yang2024qwen2}
An~Yang, Beichen Zhang, Binyuan Hui, Bofei Gao, Bowen Yu, Chengpeng Li, Dayiheng Liu, Jianhong Tu, Jingren Zhou, Junyang Lin, et~al.
\newblock Qwen2.5-math technical report: Toward mathematical expert model via self-improvement.
\newblock \emph{arXiv preprint arXiv:2409.12122}, 2024.

\bibitem[Yang et~al.(2025)Yang, Li, Yang, Zhang, Hui, Zheng, Yu, Gao, Huang, Lv, et~al.]{yang2025qwen3}
An~Yang, Anfeng Li, Baosong Yang, Beichen Zhang, Binyuan Hui, Bo~Zheng, Bowen Yu, Chang Gao, Chengen Huang, Chenxu Lv, et~al.
\newblock Qwen3 technical report.
\newblock \emph{arXiv preprint arXiv:2505.09388}, 2025.

\bibitem[Ye et~al.(2026)Ye, Dang, Fang, Wang, Zhang, Lv, Zhang, Peng, Hu, and Chua]{ye2026implicit}
Hao Ye, Jisheng Dang, Junfeng Fang, Bimei Wang, Yizhou Zhang, Ning Lv, Wencan Zhang, Hong Peng, Bin Hu, and Tat-Seng Chua.
\newblock On the implicit reward overfitting and the low-rank dynamics in {RLVR}.
\newblock \emph{arXiv preprint arXiv:2605.06523}, 2026.

\bibitem[Yue et~al.(2025)Yue, Chen, Lu, Zhao, Wang, Yue, Song, and Huang]{yue2025does}
Yang Yue, Zhiqi Chen, Rui Lu, Andrew Zhao, Zhaokai Wang, Yang Yue, Shiji Song, and Gao Huang.
\newblock Does reinforcement learning really incentivize reasoning capacity in {LLM}s beyond the base model?
\newblock In \emph{The Thirty-ninth Annual Conference on Neural Information Processing Systems}, 2025.

\bibitem[Zeng et~al.(2025)Zeng, Yin, Xu, and Liu]{zeng2025generative}
Boya Zeng, Yida Yin, Zhiqiu Xu, and Zhuang Liu.
\newblock Generative modeling of weights: Generalization or memorization?
\newblock \emph{arXiv preprint arXiv:2506.07998}, 2025.

\bibitem[Zheng et~al.(2025)Zheng, Wang, Ji, Huang, and Peng]{zheng2025expo}
Chujie Zheng, Ziqi Wang, Heng Ji, Minlie Huang, and Nanyun Peng.
\newblock Model extrapolation expedites alignment.
\newblock In \emph{Proceedings of the 63rd Annual Meeting of the Association for Computational Linguistics (Volume 1: Long Papers)}, pages 1025--1041, 2025.

\bibitem[Zhu et~al.(2025{\natexlab{a}})Zhu, Zhang, Huang, Su, Liu, Zhao, Fedorov, Pirsiavash, Sha, Lee, et~al.]{zhu2025rlvr}
Hanqing Zhu, Zhenyu Zhang, Hanxian Huang, DiJia Su, Zechun Liu, Jiawei Zhao, Igor Fedorov, Hamed Pirsiavash, Zhizhou Sha, Jinwon Lee, et~al.
\newblock The path not taken: {RLVR} provably learns off the principals.
\newblock \emph{arXiv preprint arXiv:2511.08567}, 2025{\natexlab{a}}.

\bibitem[Zhu et~al.(2025{\natexlab{b}})Zhu, Xia, Wei, Chen, Chen, and Meng]{zhu2025the}
Xinyu Zhu, Mengzhou Xia, Zhepei Wei, Wei-Lin Chen, Danqi Chen, and Yu~Meng.
\newblock The surprising effectiveness of negative reinforcement in {LLM} reasoning.
\newblock In \emph{The Thirty-ninth Annual Conference on Neural Information Processing Systems}, 2025{\natexlab{b}}.

\end{thebibliography}
